\documentclass[11pt,a4paper]{article} 
\usepackage{geometry}
\usepackage{graphicx}
\usepackage{parskip}
\usepackage{textcomp}
\usepackage{color}   
\usepackage[utf8]{inputenc}
\usepackage{authblk}
\usepackage{amsmath}
\usepackage{color}
\usepackage{epsfig}

\date{}

\title{Entropy analysis of word-length series of natural language texts: Effects of text language and genre\thanks{Preprint of an article published in International Journal of Bifurcation and Chaos, 22, 1250223, (2012) DOI: 10.1142/S0218127412502239 © copyright World Scientific Publishing Company [Journal URL:http://www.worldscientific.com/worldscinet/ijbc]}}

\author[1]{Maria Kalimeri}
\author[2]{Vassilios Constantoudis}
\author[1]{Constantinos Papadimitriou}
\author[1]{Kostantinos Karamanos}
\author[1]{Fotis K. Diakonos}
\author[3]{Haris Papageorgiou}
\affil[1]{Department of Physics, University of Athens, GR-15784, Greece}
\affil[2]{Department of Microelectronics, NCSR Demokritos, Aghia Paraskevi, Athens 15310, Greece}
\affil[3]{Institute for Language and Speech Processing, Athens, Greece}

\begin{document}
\maketitle

\begin{abstract}
We estimate the $n$-gram entropies of natural language texts in word-length representation and find that these are sensitive to text language and genre. We attribute this sensitivity to changes in the probability distribution of the lengths of single words and emphasize the crucial role of the uniformity of probabilities of having words with length between five and ten. Furthermore, comparison with the entropies of shuffled data reveals the impact of word length correlations on the estimated $n$-gram entropies.
\end{abstract}


\section{Introduction}
The output of many physical, biological and social processes can be recorded in the form of time series. Most times, these time series are neither totally random nor completely periodic but lie in between these two limits exhibiting correlations in a limited range. Several methods have been proposed for the quantification of the degree of randomness (or inversely the information content) in a time series. 

One of the most popular is the estimation of the properly defined entropy \cite{Shan1}, \cite{Khin}, \cite{Ebel1}, \cite{Ebel2}, \cite{Nico1}, \cite{Kara1}, \cite{Kara2}.  The entropy can be calculated either directly from the time series itself or after its transformation to a sequence of symbols. This transformation is usually performed by means of a partition of the state space of the time series variable. However, in the case of discrete time series (time series with discrete variable values) and especially when the range of variable values is limited, this transformation is straightforward and the time series itself can be considered a symbolic sequence with alphabet the range of variable values appeared in time series.  

A text written in natural language can also be represented as a time series. The ``time'' here is the position of the word in the text and the recorded variable is a word property such as the length or the frequency of its appearance in the text, or the rank according to its frequency of appearance, or the sum of the Unicode values of the letters of a word \cite{Shan2}, \cite{Ausl}, \cite{Kosm}, \cite{Sahi}, \cite{Mont}. In all cases, the variable takes on discrete values and thus the time series is actually a symbolic sequence with alphabet depending on the used word property. 
The degree of randomness in a symbolic sequence can be characterized by the block entropies (or $n$-gram entropies as refer to here) which generalize the Shannon entropy to $n$-grams i.e. blocks with $n$ symbols \cite{Nico1}, \cite{Ebel1}.

In this work, we employ the word-length series representation of natural language texts and estimate the $n$-gram (unigram, bigram and trigram) entropies defined in this representation.  The aim is to investigate the sensitivity of these entropies to text language (Greek or English), genre (news (politics, economy, sports) or literature) and word length correlations (comparison with shuffled series).

In our previous work \cite{Papa}, we investigated the Shannon and Kolmogorov-Sinai entropies of binary symbolic sequences generated by natural language written texts and demonstrated their sensitivity to the text language and genre. Here, we study the texts of the same corpus in a different representation as word length series and estimate their $n$-gram entropies in order to examine more straightforwardly the role of the distributions and correlations of word lengths in the effects of text language and genre. 
The paper structures as follows: In Section 2 we describe the corpus of texts analyzed, while in Section 3 we provide the methodology used for the representation of texts and the estimation of $n$-gram entropies. Section 4 presents and discusses the results of our analysis. In particular, first we show the effects of text language and genre on $n$-gram entropies and then interpret these in terms of the normalized frequency distributions. Finally, we compare these entropy values with the entropies of the shuffled data to identify the contribution of unigram (single word) distribution to bigram and trigram entropies and this way to reveal the effects of word position correlation on $n$-gram entropies. Section 5 summarizes the main findings of the paper.

\section{Corpus description}
The corpus we analyze comprises texts written in two languages: English and modern Greek and belongs to two different genres: literary works and news articles on the Web. The whole corpus accounts for a total of 5.4M words. (see Table 1 for a more detailed breakdown of corpus into languages and genres)

\begin{table}[h]
\begin{center}
\caption{Corpus breakdown into languages and genres.}
{\begin{tabular}{l c c c}\\[-2pt]
    \hline 
{} & Number of English words	& Number of Greek words	& Total Number of words\\[6pt]
\hline\\[-2pt]
Web - Politics  &0.84M     &0.76M             &1.6M \\[2pt]
Web - Economy   &0.84M     &0.74M             &1.4M \\[2pt]
Web - Sports    &0.89M     &0.79M             &1.7M \\[2pt]
Literary works  &0.42M     &0.25M             &0.7M \\[2pt]
Total           &2.99M     &2.54M             &5.4M \\[1pt]
    \hline 
\end{tabular}}
\end{center}
\end{table}

The Web dataset spans a variety of topics made between January 1st and March 31st, 2008 in three categories: sports, politics and economy and includes texts written by different authors. This dataset was derived from major (English and Greek) news sites by exploiting a focused Web Crawler and an information extraction platform where various processes have taken place (link extraction, html cleansing, URL canonicalization, utf8 encoding, metadata extraction).The dataset includes the text as indicated, as well as metadata such as date/time of posting, date/time of editing, author name, title of the article, URL link, tags/categories. The metadata is formatted in XML following a specific XML Schema. 

The literary dataset comprises text samples chosen from either literary e-books (English) or books which are part of the Hellenic National corpus\footnote{http://hnc.ilsp.gr/en/}.

\section{Methodology}
We construct the time series from the corpus data by mapping each document to a sequence of numbers where every number represents the length of the respective word. The resultant sequence consists of integers with a minimum equal to 1 and a maximum equal to the length of the longest word in the specific document. An example of this representation is shown in Figure 1.

\begin{figure}[ht]
\begin{center}
\epsfig{file=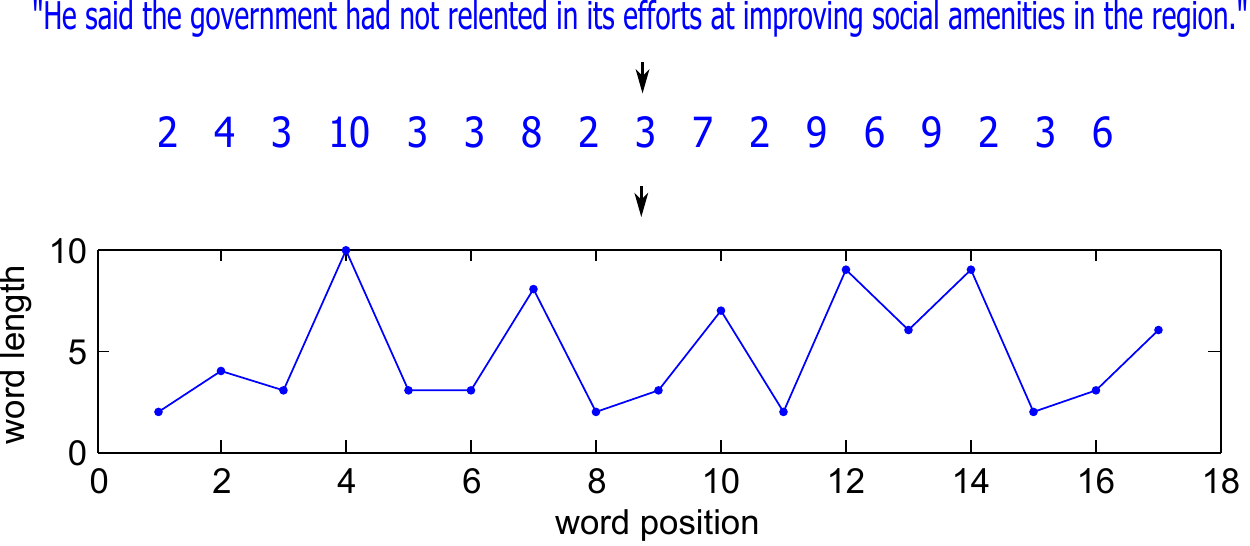,width=4in} 
\end{center}
\caption{The mapping procedure of a text to a discrete time series.}
\label{fig1}
\end{figure}

Then the sequences (discrete time series) of all documents belonging to the same genre (e.g. all political documents) are concatenated and finally one sequence per genre is obtained. The latter incorporates the author diversity, web source or book variation and topic variety of the specific genre. 

We focus our interest on the estimation of the $n$-gram entropies of the above mentioned sequences. Our $n$-gram entropies are block entropies that extend Shannon's classical definition of the entropy of a single state (unigram) to the entropy of a succession of states ($n$-grams) \cite{Nico1}, \cite{Kara1}. Actually, they account for the form of the probability distribution of $n$-grams (blocks) and large entropy values imply more uniform distributions of the appearance probabilities of $n$-grams (blocks). 

In the general case, following the ``gliding'' reading procedure, the calculation of $n$-gram entropies is done as follows \cite{Nico2}, \cite{Nico3}, \cite{Nico4}. \\
Let $S = (s_1,s_2,...,s_N)$, be a sequence of length $N$  where $s_i$  are symbols taken from a finite size alphabet. We set $K=N-n+1$ and define the $n$-grams $n_j=(s_j,...,s_{j+n-1})$ where $j=1,2,...,K$ (Figure 2).

\begin{figure}[h!]
\begin{center}
\centerline{\includegraphics[width=0.6\textwidth]{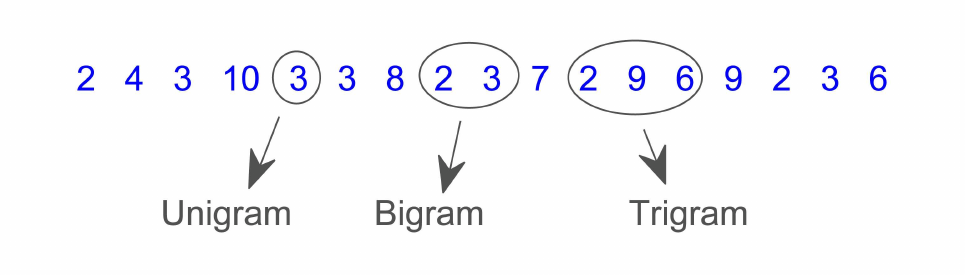}}
\end{center}
\caption{Example of unigram, bigram and trigram.}
\label{fig2}
\end{figure}

The (Shannon-like) $n$-gram entropy is defined as 
\begin{equation}
\Phi_n=-\sum_{(s_1,...,s_n)}p^{(n)}(s_1,...,s_n)\ln(p^{(n)}(s_1,...s_n))\,\,
.\label{this}
\end{equation}
where the summation is over all unique $n$-grams and the probability of occurrence of a unique $n$-gram $(s_1,...,s_n)$, denoted by $p^{(n)}(s_1,...,s_n)$ is defined by the fraction  
\begin{center}
$\frac{\text{No. of $n$-grams }(s_1,...,s_n) \text{ encountered when gliding}}{\text{Total No. of $n$-grams}}\,\,$
\end{center}
This quantity is maximized for the so called ``normal sequence'' where all possible $n$-grams appear with equal probability \cite{Borw2} \cite{Mann}.

We now return to our word-length sequences. In order to, both, retain the short document character of the series and have valid statistics, we go further into fragmentizing them in segments of $N=1000$  words. The entropy calculation is done over each one of these segments and the final value results from averaging over all segments of the same genre.

\section{Results and Discussion}
\subsection{Effects of text language and genre on $n$-gram Entropies}
In Figure 3 we present the results of the estimation of the $n$-gram entropies $\Phi_{1}$, $\Phi_{2}$ and $\Phi_{3}$ for both languages and all genres.  As said in Section 3, the shown values are averages over the entropies of all segments of the same language and genre series. 

\begin{figure}[ht]
\begin{center}
\psfig{file=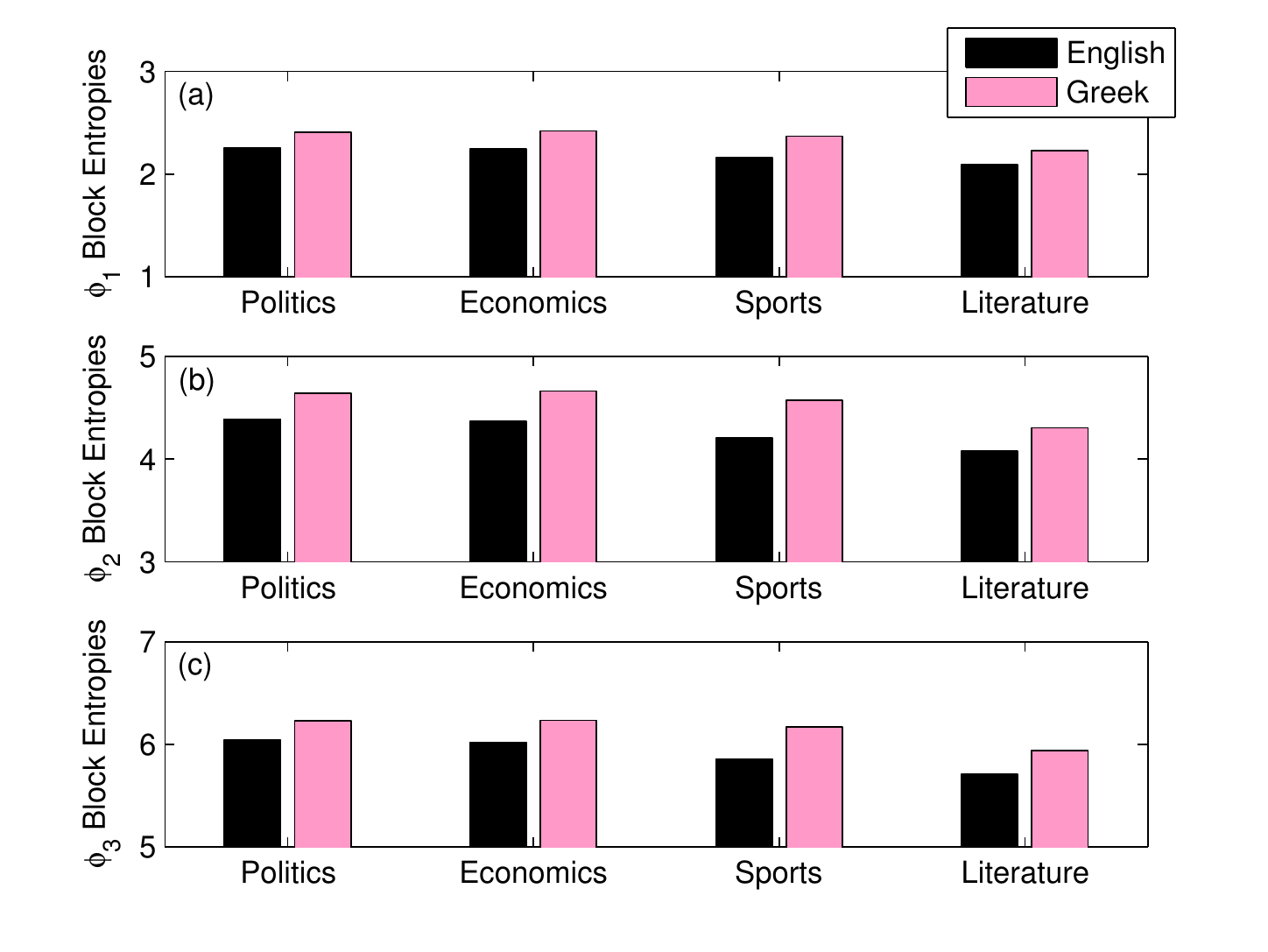,width=4in} 
\end{center}
\caption{Bar plots showing unigram $\Phi_{1}$ in (a), bigram $\Phi_{2}$ in (b) and trigram $\Phi_{3}$ in (c), for all text genres in both languages with $N=1000$ and averaging over all segments.}
\label{fig3}
\end{figure}

One can notice that, in all genres, the $n$-gram entropies $\Phi_{1}$, $\Phi_{2}$ and $\Phi_{3}$ of the Greek texts take higher values than the English ones. Regarding the differences among text genres, we have found that politics and economy texts exhibit the higher $n$-gram entropies followed by sports, while the literature works are characterized by the smallest entropies.   

To investigate the dependence of this result on segment length $N$, we have repeated the calculations for a wide range of segment lengths ($250 \leq N \leq 3000$ words). In Figure 4, we show the results for the bigram (two-word) entropy $\Phi_{2}$ of all text genres and languages. Similar behavior has been obtained for $\Phi_{1}$ and $\Phi_{3}$. We can see that although the entropy values are sensitive to the choice of $N$, their order with respect to language and genre remains unaltered and similar to what we have found for $N=1000$ (see Figure 3).  

\begin{figure}[ht]
\begin{center}
\psfig{file=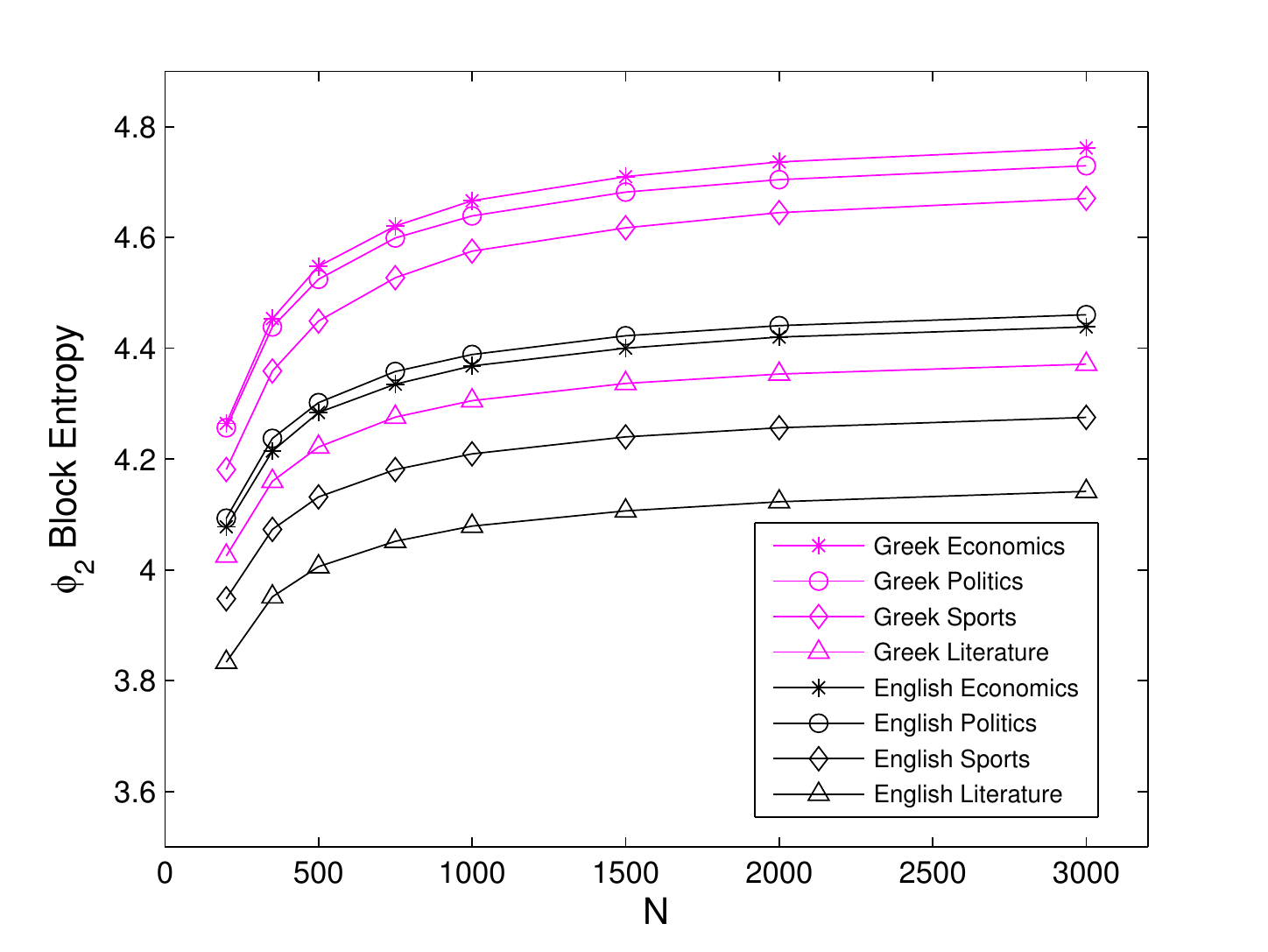,width=4in} 
\end{center}
\caption{$\Phi_{2}$ versus segment length $N$ for all genres and languages. Notice the dependence of the entropy values on $N$ and the independence of their order w.r.t. text language and genre.}
\label{fig4}
\end{figure}

\subsection{Distributions of $n$-gram frequencies}
What are the origins of these differences in entropy values between different languages and genres? It is well-known that entropy is a measure of the uniformity of the probability (normalized frequency of appearance) distribution of $n$-grams. Thus, we expect texts with higher entropy to have a more uniform $n$-gram distribution. Indeed, Figure 5 shows the probability distribution of unigrams (Figure 5(a)), bigrams  (Figure 5(b)) and trigrams (Figure 5(c)) of politics genre for both languages versus the rank of the $n$-gram, which is defined w.r.t. its probability (1 is for the most frequently appeared $n$-gram, 2 for the second one and so on).
 
\begin{figure}[ht]
\begin{center}
\psfig{file=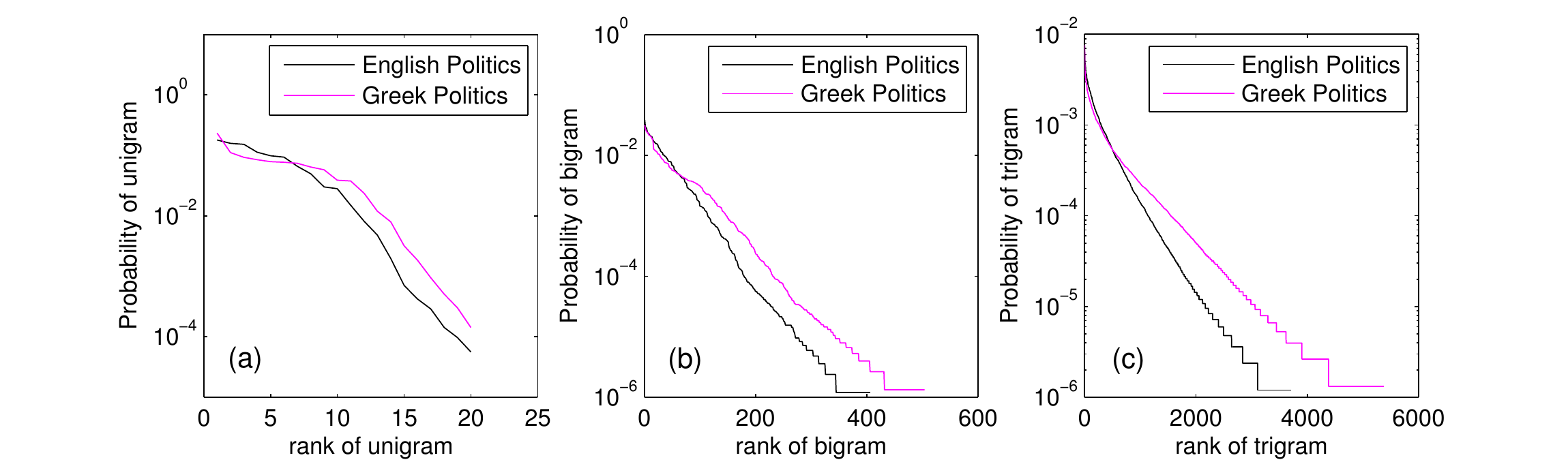,width=6.5in} 
\end{center}
\caption{Probabilities vs $n$-gram rank for unigrams (a), bigrams (b) and trigrams (c) of politics' texts for both languages.}
\label{fig5}
\end{figure}

$\Phi_{1}$ exhibits a plateau or slow decrease at low ranks and an almost exponential decrease at high ranks. The plateau is wider and more pronounced in the Greek texts and this gives rise to their increased unigram entropy value. On the other hand, the exponential decrease has similar rates in both languages. In order to explain better the behavior of $\Phi_{1}$ in terms of word lengths, we present in different diagrams the probability distributions versus the word length of all genres for both languages (Figure 6).

\begin{figure}[ht]
\begin{center}
\psfig{file=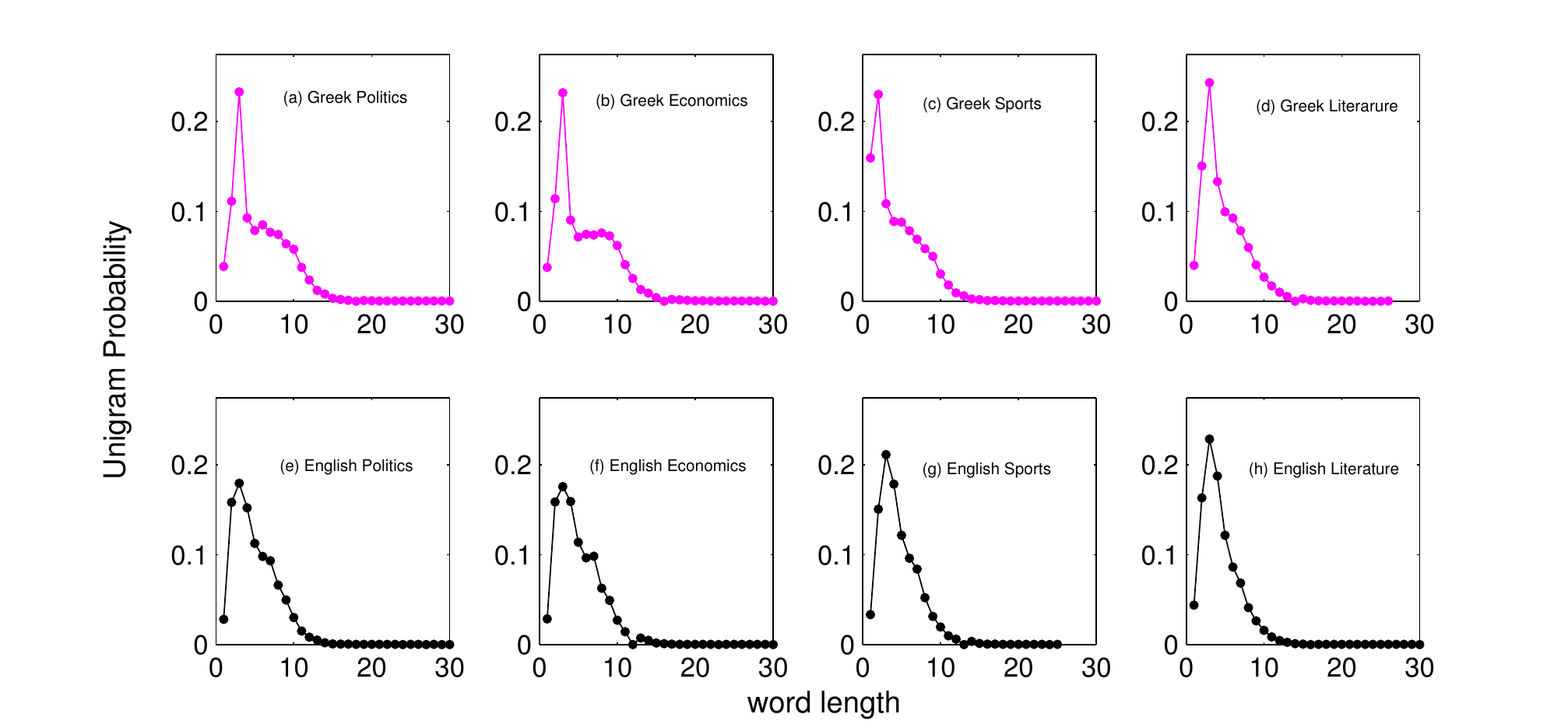,width=6.5in} 
\end{center}
\caption{Unigram probability distributions versus word length for both languages and all genres analyzed in the paper.}
\label{fig6}
\end{figure}

The distributions in Figure 6 show that the main contribution to the differences in $\Phi_{1}$ between different languages and genres comes from the probabilities of the words with length $~5-10$. It seems that, for the same genre, the distribution of these probabilities in the Greek texts is more uniform than in the English ones. Regarding the text genre, this approximate plateau is suppressed to smaller word lengths as we move from the economy-politics texts to sports and then to literature and this suppression causes the differences among the unigram entropies of various genres.
 
In Figure 5(b), we can see then in $\Phi_{2}$ a plateau is only present in Greek bigrams at middle ranks, while the exponential drop off at high ranks is now marginally slower in Greek texts. This difference means that the probability in the Greek texts to meet bigrams with word lengths $~5-10$ is more uniform that in the English texts of the same genre and explains the higher Greek $\Phi_{2}$ values.  

Finally, the plateau is found to be almost absent in both Greek and English $\Phi_{3}$ behavior. Actually, one can notice a super-exponential drop off at low ranks following with an almost exponential decrease at high ranks. The difference of two languages comes from the faster (slower) decrease at low (high) ranks of Greek $\Phi_{3}$ value. 

A common characteristic of the dependencies of all $n$-gram entropies on $n$-gram rank is the crossing of Greek with English curves at the region of $n$-grams with word lengths $~5-10$.  

The $n$-gram probability distributions can be also used for a deeper understanding of the entropy differences among text genres. We can distinguish two extreme cases. When a plateau is present, then the difference among genres comes mainly from the width of plateau. One example is shown in Figure 7(a), where we can see that the plateau is wider in politics/economy texts, a bit narrower in sports and marginally present in literature texts. The second case occurs when no plateau exists. Here, the difference comes mainly from the rate of decrease at all ranks. As we move from the politics-economy genre to sports and literature the rate increases making the distribution less uniform and the $n$-gram entropy smaller (see  Figure 7(b) for the $\Phi_{3}$ of English texts).

\begin{figure}[ht]
\begin{center}
\psfig{file=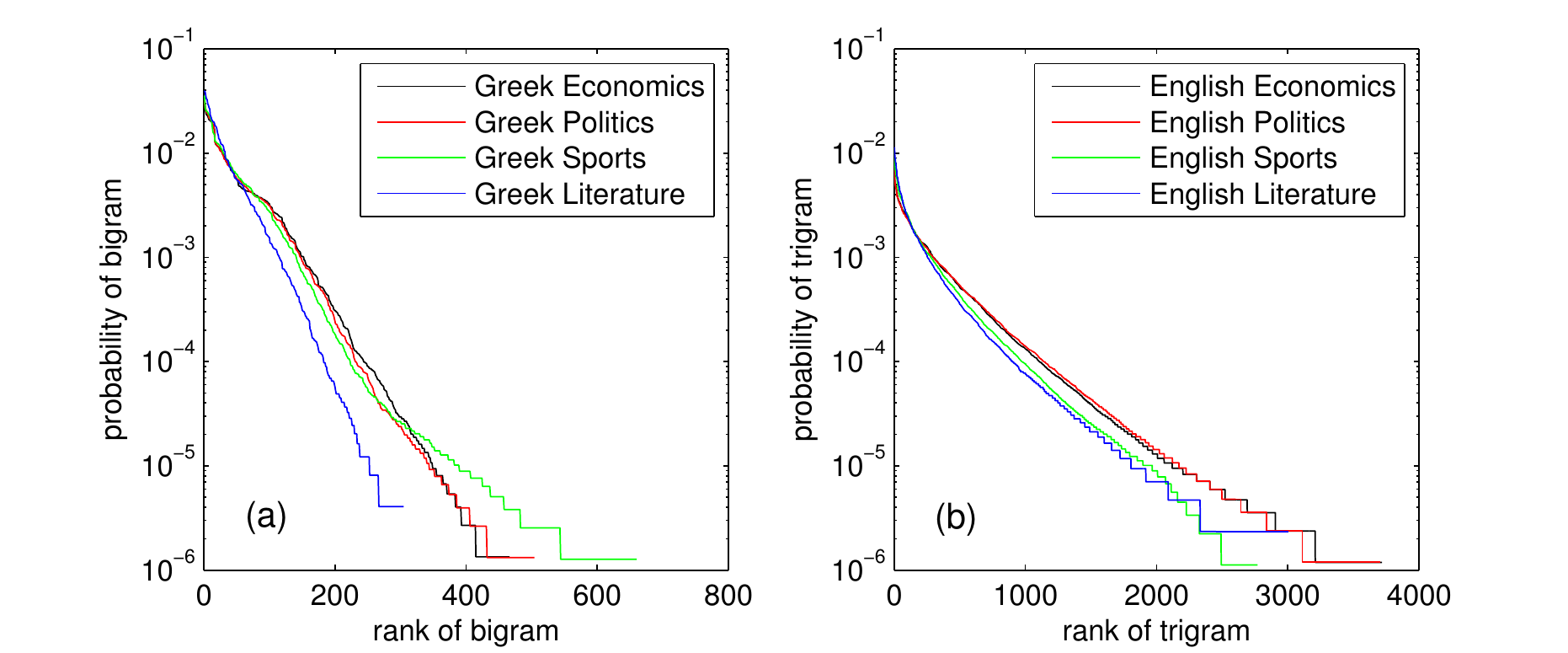,width=6in} 
\end{center}
\caption{Probability distributions of bigrams (two-word blocks) in Greek data (a) and trigrams (three-word blocks) in English data (b). Notice that the main contribution to differences originates from the plateau at middle ranks in (a) and the decreasing rate at low and high ranks in (b).}
\label{fig7}
\end{figure}

\subsection{Comparison with shuffled data}
The unigram entropy $\Phi_{1}$ is a measure of the uniformity of the probability distribution of single words. No correlations between the word lengths at different positions are accounted for. On the other hand, $\Phi_{2}$ and $\Phi_{3}$ by definition contain information about the correlations between the lengths of two and three adjacent words respectively. However, at the same time $\Phi_{2}$ and $\Phi_{3}$ are affected by the unigram probability distribution quantified by $\Phi_{1}$. Thus, one can say that we have two sources for differentiations in $\Phi_{2}$ and $\Phi_{3}$ values: the inter-word-length correlations and the unigram probability distributions. Which is more crucial in the differences of $\Phi_{2}$ and $\Phi_{3}$ among different languages and genres found above? To address this question, we have estimated the $n$-gram entropies of shuffled series having the same unigram distributions to the real data for all categories analyzed in the previous sub-sections. The results are shown in Table 2.

\begin{table}[ht]
\begin{center}
\caption{Bigram ($\Phi_{2}$) and trigram ($\Phi_{3}$) entropies for the real and the shuffled data.}
{\begin{tabular}{l c c c c c c c c}\\[-2pt]
\hline
{} & PolE	&PolG	&EcoE	&EcoG &SpoE	&SpoG	&LitE	&LitG\\[6pt]
\hline\\[-2pt]
$\Phi_{2}$             &4.389 &4.639 &4.366	&4.664	&4.209	&4.575	&4.079	&4.306\\[1pt]
$\Phi_{2}$(shuffled)   &4.438	&4.707 &4.416	&4.735	&4.254	&4.650	&4.149	&4.375\\[2pt]
$\Phi_{3}$             &6.041	&6.227 &6.018	&6.232	&5.856	&6.166	&5.708	&5.940\\[2pt]
$\Phi_{3}$(shuffled)   &6.115	&6.301 &6.095	&6.320	&5.923	&6.262	&5.807	&6.021\\[1pt]
\hline
\end{tabular}}
\end{center}
\end{table}

The first  finding obtained by the comparison of the shuffled with the real $n$-gram entropies is that the major contribution to bigram and trigram entropies comes from the unigram probability distributions, since the values of the shuffled series are very close to those of real texts. Thus, the differences in the single-word length distributions and hence $\Phi_{1}$ are responsible for the effects of text language and genre on $n$-gram entropies. 

However, a closer inspection reveals that the shuffled word-length series have in all cases slightly higher entropies than the real ones. This difference demonstrates that the entropies are slightly sensitive to the correlations between word positions in the word-length representation of texts. The entropy effects of the word length correlations seem to be larger in Greek texts especially in $\Phi_{2}$, although the differences are slight to obtain conclusions with certainty.  

\section{Conclusions}
In this paper, we have estimated the $n$-gram entropies of natural language written texts in the word-length series representation and examined their sensitivity to the language and the genre of texts. We have found that at least the unigram, bigram and trigram entropies are sensitive to both text language and genre. Greek texts exhibit higher entropies than English ones, while web news texts with politics and economy content have higher entropies than sports news and literature works which are characterized by the lowest values of $n$-gram entropies.

The entropy differences can be explained in the terms of the probability distributions of single words w.r.t. their length. More specifically the main contribution comes from the probability of meeting words with length between 5 and an upper limit that depends on the text genre. For texts of the same genre type, the frequencies of these word lengths are more uniform in Greek than in English ones and exhibit a more pronounced plateau. For texts of the same language, the genre type affects the length range of the words with almost similar probability. As we move from the economy/politics texts to sports and literature texts, the range is suppressed and its upper limit approaches lower lengths. 

Furthermore, it has been shown that the differences in unigram distributions provide the main contribution to the behavior of $\Phi_{2}$ and $\Phi_{3}$ w.r.t. text language and genre. However, shuffled data with the same single word distributions with the real series have systematically little higher $\Phi_{2}$ and $\Phi_{3}$ values. These entropy differences with the real texts may be considered that quantify the short-range correlations between adjacent words in the word-length representation of a text.

\bibliographystyle{ieeetr}

\end{document}